\documentclass[journal]{IEEEtran}

\ifCLASSINFOpdf
\else
   \usepackage[dvips]{graphicx}
\fi
\usepackage{url}

\usepackage{graphicx}
\usepackage{color,xcolor}
\usepackage{bbding}

\usepackage{algorithmic}
\usepackage{amsmath,amssymb,amsfonts}
\usepackage[justification=centering]{caption}
\begin{document}

\title{Modal-Adaptive Gated Recoding Network\protect\\ for  RGB-D Salient Object Detection}
\author{Jinchao Zhu, Xiaoyu Zhang$^{*}$,  Xian Fang,  Feng Dong, Qiu Yu
\thanks{Jinchao Zhu, Xiaoyu Zhang (corresponding author) and Qiu Yu are with the College of Artificial Intelligence, Nankai University, Tianjin, China.
        (e-mail:jczhu@mail.nankai.edu.cn)}
\thanks{Feng Dong is with the School of Finance, Tianjin University of Finance and Economics, Tianjin, China.
        (e-mail:dongfengdeyx@163.cn)}
\thanks{Xian Fang is with the College of Computer Science, Nankai University, Tianjin, China.
        (e-mail:xianfang@mail.nankai.edu.cn)}
        }

\markboth{Journal of \LaTeX\ Class Files, Vol. 14, No. 8, August 2015}
{Shell \MakeLowercase{\textit{et al.}}: Bare Demo of IEEEtran.cls for IEEE Journals}
\maketitle

\begin{abstract}
The multi-modal salient object detection model based on RGB-D information has better robustness in the real world.
However, it remains nontrivial to better adaptively balance effective multi-modal information in the feature fusion phase.
In this letter, we propose a novel gated recoding network (GRNet) to evaluate the information validity of the two modes, and balance their influence.
Our framework is divided into three phases: perception phase, recoding mixing phase and feature integration phase.
First, A perception encoder is adopted to extract multi-level single-modal features, which lays the foundation for multi-modal semantic comparative analysis.
Then, a modal-adaptive gate unit (MGU) is proposed to suppress the invalid information and transfer the effective modal features to the recoding mixer and the hybrid branch decoder.
The recoding mixer is responsible for recoding and mixing the balanced multi-modal information.
Finally, the hybrid branch decoder completes the multi-level feature integration under the guidance of an optional edge guidance stream (OEGS).
Experiments and analysis on eight popular benchmarks verify that our framework performs favorably against 9 state-of-art methods.
\end{abstract}


\IEEEpeerreviewmaketitle

\section{Introduction}
Salient object detection (SOD)~\cite{2019-SPL-EACNNet}\cite{2018-SPL-BGFANet}\cite{2016-SPL-DCMC} aims to identify object-level regions with strong visual impact that attract the attention of the human visual system most in an image~\cite{what-is-SOD1}\cite{2009-CVPR-Fm}. 
With the development of deep learning, deep features have replaced hand-crafted features~\cite{2014-ICIMCS-DES}\cite{2016-SPL-DCMC}\cite{2017-ICCVW-CDCP} as a powerful tool for SOD, and promote it to better serve application-oriented tasks~\cite{2010-ACMTOG-RepFinder}\cite{2009-ACMTOG-Sketch2Photo}\cite{2009-CVPR-tracking-SBDT}.

The existing RGB-D SOD methods have proposed a variety of multi-level feature and multi-modal feature fusion techniques.
\cite{2020-TIP-DPANet} uses Ostu algorithm~\cite{1979-TSMC-Ostu} to design supervised gate modules to suppress invalid depth information.
\cite{2020-ECCV-GateNet} adopts the gated mechanism to suppress different level features in different degrees.
\cite{2020-TIP-DPANet} proposes a three-stream structure and uses an attention mechanism to better fuse multi-modal features.
However, there is still room for improvement.
Different from the above methods, we provide more structure attempts for the gate unit and propose a four-stream structure with a dual recoding mixer.
Besides, our gate unit does not need to specially design supervision labels like~\cite{2020-TIP-DPANet}.
Let's introduce our network in order of phases.



In this letter, from the perspective of multi-modal information balance, we propose a novel saliency detection method to suppress the invalid modal information and recode the balanced features.
The whole network is a 4-stream structure and includes three phases: perception phase, recoding mixing phase, and feature integration phase.
In the first phase, as shown in Fig.\ref{pipline}, the perception encoder consists of two parts (Encoder-A and Encoder-B), which are only responsible for processing single-modal data and providing multi-level features for the two-phase recoding mixer.
MGU is the bridge between perception encoder, recoding mixer, and hybrid branch decoder.
It senses and analyzes the corresponding multi-level features of the two modes, suppresses the inaccurate features, and transfers the balanced features to the recoding mixer.
In the second phase, the recoding mixer (Mixer-A and Mixer-B) recodes the balanced multi-level multi-modal features in the way of step-by-step insertion.
In the third phase, the hybrid branch decoder uses three classical feature integration structures for reference to further fuse the multi-level features provided by the mixer.
The accurate low-level detail features in the encoder are used for cross-phase multi-modal edge guidance under the regulation of OEGS.

Our contributions are summarized as follows:
\begin{itemize}
\item We propose MGU to compare and analyze the multi-level features obtained by two single-modal encoders to complete the feature validity evaluation.
    According to the evaluation results, MGU suppresses negative modal information to achieve multi-modal balance.
\item We adopt a recoding mixer to recode the new features balanced by MGU to obtain more accurate multi-level features.
The hybrid branch decoder integrates multi-level features under the guidance of OEGS.
\item Sufficient experiments conducted on 8 RGB-D SOD datasets demonstrate that the proposed method outperforms 26 state-of-the-art methods.
\end{itemize}

\section{The Proposed Method}
The motivation for the proposed network is two-fold.
First, RGB images are vulnerable to the interference of light and clutter background, resulting in inaccurate segmentation.
The depth maps are insensitive to the detail texture information in the plane region, and the semantic information provided by the depth maps is sometimes invalid.
Therefore, we seek a solution (MGU) to perceive and suppress the inaccurate modal information and then fuse the balanced multi-modal features.

Second, the edge details of RGB and depth maps are effective to optimize the edge of segmentation results, but some edge information may be misleading, so we design a decoder with optional edge guidance to better perform the multi-modal edge guidance.
\begin{figure*}[htb]
\centering

 \includegraphics[width=2\columnwidth]{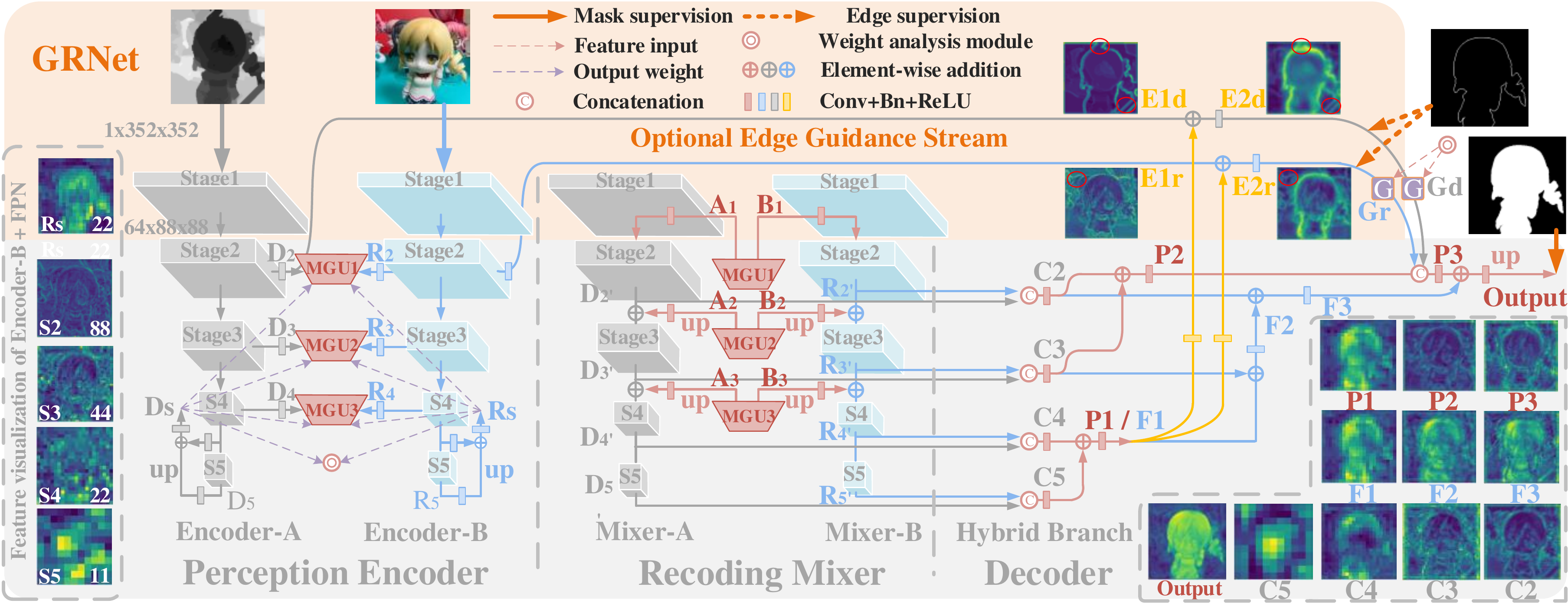} 
\caption{The overall architecture of GRNet.
The network contains three parts: single-modal Perception Encoder, multi-modal Recoding Mixer, and Hybrid Branch Decoder.
The modal-adaptive gate unit is the bridge between encoder and mixer.}
\label{pipline}
\end{figure*}

The proposed method is illustrated in Fig.\ref{pipline}, which consists of three parts: Perception Encoder, Recoding Mixer, and Hybrid Branch Decoder.
MGU and OEGS connect them.
\subsection{Perception Encoder and Modal-Adaptive Gate Unit}  
Perception Encoder is responsible for multi-level feature extraction of single-modal data and effectiveness evaluation of multi-level features.
It consists of two parts: Encoder-A and Encoder-B.
They are both made up of ResNet-50 and their output feature of each stage is processed by a convolution operation to unify the channel as 64. As shown in Fig.\ref{pipline} encoder, the final output features are $D_{2},D_{3},D_{4},D_{5},R_{2},R_{3},R_{4},R_{5}$.

Modal-adaptive gate unit (MGU) adopts multi-level features provided by Perception Encoder to evaluate the effectiveness of modal information level by level, as shown in Fig.\ref{pipline} MGU1, MGU2, MGU3.
Taking MGU1 as an example, its inputs are depth semantic feature ($Ds$, purple arrow), RGB semantic feature ($Rs$, purple arrow), current level depth feature ($D_{2}$, grey arrow), and current level RGB feature ($R_{2}$, blue arrow).
Semantic features ($Ds$,$Rs$) are obtained by integrating the features of the deepest two stages (4,5) through FPN structure~\cite{2017-CVPR-FPN}.
The deeper the feature, the larger the receptive field and the richer the semantic information.  
Because the feature size of the 5th stage is too small (11x11), it will lose the spatial structure information, so we combine it with the 4th stage feature (22x22). The visualization in the dotted box on the left side of Fig.\ref{pipline} confirms the above statement.

\begin{figure}[t]
\centering
 \includegraphics[width=1\columnwidth]{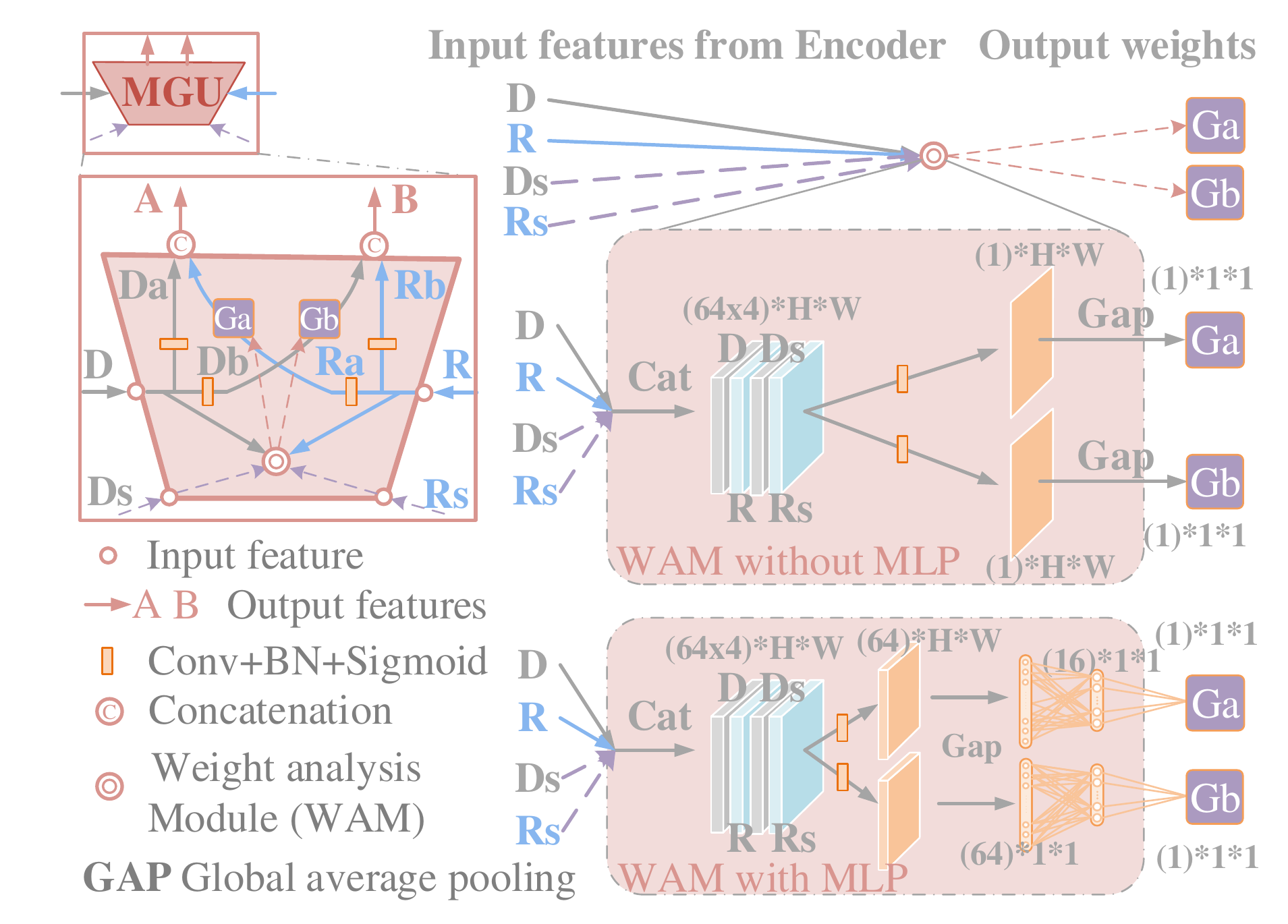} 
\caption{The architecture details of the modal-adaptive gate unit (MGU).
The input features of the MGU are from the Encoder and the output features are fed into the Mixer.
Weight analysis modules (WAM) are shown in the right.
}
\label{MGU}
\end{figure}

Fig.\ref{MGU} shows the details of the MGU, the input features ($Ds$,$Rs$,$D_{2}$,$R_{2}$) are sent to the weight analysis module (WAM).
We analyze the details of the internal WAM of MGU in detail on the right.
Here, we propose two kinds of weight analysis schemes with and without multi-layer perceptron (MLP)~\cite{2018-arXiv-BAM}.
Taking the calculation of $G_{a}$ as an example, the calculation process of the scheme without MLP is as follows:
\begin{small}
	\begin{align}\label{mgu}
        G_{a} = Gap(C_{S}^{1}(Cat(D_{2},R_{2},Ds,Rs))).  
	\end{align}
\end{small}
The process of the scheme with MLP is as follows:
\begin{small}
	\begin{align}\label{mgu}
        G_{a} = MLP_{S}^{64-16-1}(Gap(C_{S}^{64}(Cat(D_{2},R_{2},Ds,Rs)))),
	\end{align}
\end{small}
where 64 in $C_{S}^{64}$ means that the output feature is 64 channels,
$_{R}$ and $_{S}$ mean that the activation function is ReLU and Sigmoid respectively.
$Gap$ and $Cat$ are global average pooling and concatenation.
$MLP_{S}^{64-16-1}$ means that the hidden and output layers of multi-layer perceptron are 64 and 1.
After getting the balance weights ($G_{a}$,$G_{b}$) through WAM, we use them to suppress the invalid features to obtain the modal-adaptive fusion features $A$ and $B$.
The process is as follows:
\begin{small}
	\begin{align}\label{mgu}
        A = Cat(C^{64}_{S}(D_{2})+G_{a}\times(C^{64}_{S}(R_{2}))),\\
        B = Cat(C^{64}_{S}(D_{2})\times G_{b}+(C^{64}_{S}(R_{2}))),
	\end{align}
\end{small}
where $\times$ and means element-wise multiplication.

\begin{table*}[htbp]
  \centering
  \caption{Quatitative comparision.
  $t$, $e$ and $n$ represent the traditional method, edge guidance method, and RGB method.}
  \resizebox{\textwidth}{!}{
    \begin{tabular}{|l|cc|cc|cc|cc|cc|cc|cc|cc|}
    \hline
    \multicolumn{1}{|c|}{Model} & \multicolumn{2}{c|}{LFSD}             & \multicolumn{2}{c|}{NJUD}             & \multicolumn{2}{c|}{NLPR}             & \multicolumn{2}{c|}{RGBD135}          & \multicolumn{2}{c|}{SIP}              & \multicolumn{2}{c|}{SSD}              & \multicolumn{2}{c|}{STEREO}           & \multicolumn{2}{c|}{DUTRGBD} \\
                      & F$^{w}_{\beta}\uparrow$                & MAE$\downarrow$               & F$^{w}_{\beta}\uparrow$                & MAE$\downarrow$                & F$^{w}_{\beta}\uparrow$                & MAE$\downarrow$                & F$^{w}_{\beta}\uparrow$                & MAE$\downarrow$                & F$^{w}_{\beta}\uparrow$                & MAE$\downarrow$                & F$^{w}_{\beta}\uparrow$                & MAE$\downarrow$                & F$^{w}_{\beta}\uparrow$                & MAE$\downarrow$                & F$^{w}_{\beta}\uparrow$                & MAE$\downarrow$  \\
    \hline
    CDCP(ICCVW17)$^{t}$    & .490              & .206              & .479              & .188              & .477              & .116              & .457              & .121              & .397              & .224              & .403              & .219              & .558              & .149              & .491              & .165 \\
    \hline
    D3Net(TNNLS20)    & .756              & .099              & .833              & .051              & .826              & .034              & .831              & .030              & .793              & .063              & .780              & .058              & .815              & .054              & -                 & - \\
    S2MA(CVPR20)      & .772              & .094              & .842              & .053              & .852              & .030              & \textcolor[rgb]{ .357,  .608,  .835}{.892} & \textcolor[rgb]{ .357,  .608,  .835}{.021} & -                 & -                 & .787              & .052              & .825              & .051              & .856              & .046 \\
    CoNet(ECCV20)     & \textcolor[rgb]{ .357,  .608,  .835}{.815} & \textcolor[rgb]{ 1,  0,  0}{.071} & .849              & .046              & .842              & .031              & .849              & .028              & .803              & .063              & .780              & .059              & -                & -                & \textcolor[rgb]{ .357,  .608,  .835}{.891} & \textcolor[rgb]{ 1,  0,  0}{.033} \\
    DPANet(TIP20)     & .814              & \textcolor[rgb]{ .357,  .608,  .835}{.072} & .882              & \textcolor[rgb]{ .357,  .608,  .835}{.035} & .875              & .024              & .868              & .023              & .833              & .051              & \textcolor[rgb]{ 1,  0,  0}{.826} & \textcolor[rgb]{ .357,  .608,  .835}{.042} & -                 & -                 & .852              & .048 \\
    CDNet(TIP21)      & .812              & .077              & .865              & .043              & .873              & .026              & .886              & .022              & .798              & .065              & .809              & .049              & .845              & .045              & .829              & .054 \\
    cmSalGAN(TMM21)   & .761              & .097              & .846              & .046              & .855              & .027              & .840              & .028              & .795              & .064              & .650              & .086              & -                 & -                 & .786              & .068 \\
    \hline
    SCRN(ICCV19)$^{e}$     & .728              & .109              & .840              & .047              & .833              & .032              & .809              & .033              & .803              & .058              & .774              & .054              & .833              & .046              & .856              & .043 \\
    F3(AAAI20)$^{n}$       & .754              & .098              & .864              & .041              & .863              & .029              & .836              & .030              & .824              & .053              & .796              & .052              & .857              & .040              & .875              & .039 \\
    \hline
    GRNet-MLP         & .802              & .079              & \textcolor[rgb]{ .357,  .608,  .835}{.886} & .036              & \textcolor[rgb]{ .357,  .608,  .835}{.887} & \textcolor[rgb]{ .357,  .608,  .835}{.023} & \textcolor[rgb]{ 1,  0,  0}{.897} & \textcolor[rgb]{ 1,  0,  0}{.020} & \textcolor[rgb]{ .357,  .608,  .835}{.840} & \textcolor[rgb]{ .357,  .608,  .835}{.049} & .822              & .044              & \textcolor[rgb]{ 1,  0,  0}{.871} & \textcolor[rgb]{ 1,  0,  0}{.036} & \textcolor[rgb]{ 1,  0,  0}{.893} & \textcolor[rgb]{ 1,  0,  0}{.033} \\
    GRNet+MLP(augmentation) & \textcolor[rgb]{ 1,  0,  0}{.820} & .074              & \textcolor[rgb]{ 1,  0,  0}{.890} & \textcolor[rgb]{ 1,  0,  0}{.034} & \textcolor[rgb]{ 1,  0,  0}{.890} & \textcolor[rgb]{ 1,  0,  0}{.022} & .890              & \textcolor[rgb]{ 1,  0,  0}{.020} & \textcolor[rgb]{ 1,  0,  0}{.847} & \textcolor[rgb]{ 1,  0,  0}{.046} & \textcolor[rgb]{ .357,  .608,  .835}{.824} & \textcolor[rgb]{ 1,  0,  0}{.041} & \textcolor[rgb]{ .357,  .608,  .835}{.865} & \textcolor[rgb]{ .357,  .608,  .835}{.038} & .886              & \textcolor[rgb]{ .357,  .608,  .835}{.036} \\
    \hline
    \end{tabular}%
    }
  \label{T-sota}%
\end{table*}%

\subsection{Multi-Modal Recoding Mixer}

The multi-modal recoding mixer is composed of two parts (Mixer-A, Mixer-B), and their backbones both are a part of ResNet-50 (without Stage1), as shown in the middle of Fig.\ref{pipline}.
The Mixer-A and Mixer-B re-encode the modal-adaptive fusion features (A, B) in the way of step-by-step insertion.
In Fig.\ref{MGU}, we can find that the depth feature is regulated in feature $B$, and the RGB feature is regulated in feature $A$.
Therefore, Mixe-B is mainly responsible for the situation that RGB features are effective and depth information is unreliable.
Mixer-A is responsible for the situation that the depth information has reference value but the RGB data is not accurate.
We take stage2 and stage3 of Mixer-B as an example to introduce the operation process of Recoding Mixer.
\begin{small}
	\begin{align}\label{mgu}
        R_{2'} &= Res^{S2}(C^{64}_{R}(B1)),\\
        R_{3'} &= Res^{S3}(C^{256}_{R}(Up(B2))+R_{2'}),
	\end{align}
\end{small}
where $Up$($\cdot$) is upsampling 2 times. $Res^{S2}$ is the operation of ResNet-50 stage 2.

\subsection{Hybrid Branch Decoder}
\begin{figure}[t]
\centering
 \includegraphics[width=1\columnwidth]{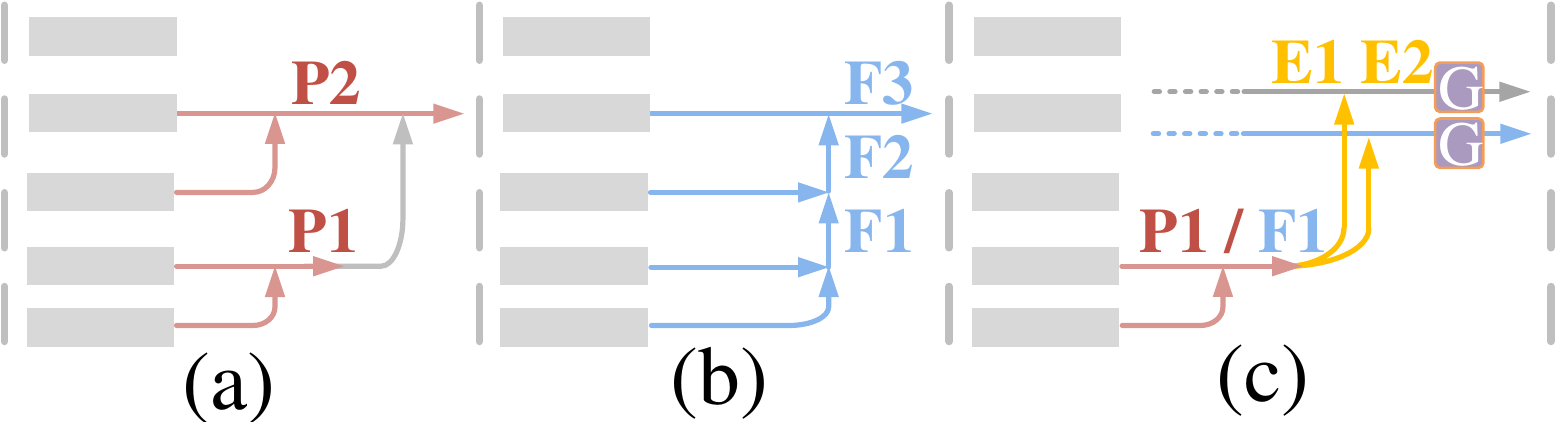} 
\caption{Analysis of multiple branches of the decoder. }
\label{decoder}
\end{figure}

The hybrid branch decoder first fuse the features ($D_{2'}$, $R_{2'}$, ..., $D_{5'}$, $R_{5'}$) of the same level output from the Recoding Mixer by concatenation operation to get features $C2$, $C3$, $C4$, $C5$.
Then three classical fusion structures (parallel structure (a), progressive structure (b), and edge guidance (c)) are adopted for multi-level feature integration.
We split the hybrid branch decoder in Fig.\ref{pipline} into three parts, as shown in Fig.\ref{decoder}.
Most existing methods either adopt parallel structure~\cite{2018-IJCAI-R3Net}\cite{2019-CVPR-PFANet}\cite{2020-ECCV-GateNet}, or progressive structure~\cite{2018-CVPR-BMPM}\cite{2018-CVPR-DGRL}\cite{2018-CVPR-PAGRN}\cite{2019-CVPR-PAGE}\cite{2019-SPL-EACNNet}\cite{2018-SPL-BGFANet}.
Besides, we also design our multi-modal optional edge guidance stream (c) based on~\cite{2019-ICCV-EGNet}.
It is worth noting that the output features after convolution in the decoder are all 64 channels.

In progressive structure, high-level features contribute more contextual guidance, which is conducive to the prediction of the main body.
However, as a residual structure, the parallel structure maximizes the value of low-level features to compensate for the details.
The P2 branch of parallel structure receives multi-modal edge guidance from OEGS, which further enhances edge optimization.
The visualization in the lower right corner of Fig.\ref{pipline} validates the above statement.

We can see that under the semantic guidance of P1/F1 (P1 is F1), the invalid edge information (the red circles of E1d, E2d, E1r, E2r in Fig.\ref{pipline}) in multi-mode edge features (E1d, E1r) is well suppressed at the semantic level.
The weight outputs (Gr, Gd) by the WAM module in the encoder further filter the two kinds of edge features at the modal level to prevent the wrong guidance of useless modal edge information.

\section{Experiments}

\subsection{Experimental Setup}
$\textit{RGB-D Salient Object Detection Datasets:}$
we follow~\cite{2019-ICCV-DMRA-DUTRGBD} to design the experiment.
For $\textbf{DUTRGBD}$~\cite{2019-ICCV-DMRA-DUTRGBD}, we utilize 800 pairs of data for training and the remaining 400 pairs for testing.
For the other datasets, we follow~\cite{2018-CVPR-PCANet}\cite{2018-PR-MMCI}\cite{2017-IROS-M3Net} to adopt 700 pairs sampled from $\textbf{NLPR}$~\cite{2014-LNCS-NLPR} and 1,485 pairs sampled from $\textbf{NJUD}$~\cite{2015-SPIC-NJUD} for training.
The remaining data are used as testing datasets: $\textbf{NLPR}$, $\textbf{NJUD}$, $\textbf{RGBD135}$~\cite{2014-ICIMCS-RGBD135}, $\textbf{SIP}$~\cite{2020-TNNLS-SINet}, $\textbf{SSD}$~\cite{2017-ICCVW-SSD}, $\textbf{STEREO}$~\cite{2012-CVPR-STEREO}, $\textbf{LSDF}$~\cite{2017-TPAMI-LFSD}.

$\textit{Implementation Details:}$ Four ResNet-50s, pre-trained on ImageNet, are used in the main part of encoders and mixers respectively.
We adopt warm-up and linear decay strategies in training.
The maximum learning rates of the backbone and other parts are set to 5e-3 and 5e-2 respectively.
SGD (stochastic gradient descent) is the optimizer.
Weight decay is 5e-4.
Momentum is 0.9.
The batch size is 16.
The training maximum epoch is set to 30.
We use a PC with RTX 2080Ti GPU and 16GB RAM for training and inference.
The input image in the test phase is set to $352\times352$.

\subsection{Evalution Merics}

We used two widely used metrics to evaluate the performance of our model and the state-of-the-art methods. 
Mean absolute error~\cite{2012-CVPR-MAE} ($\textbf{MAE}$) is adopted to estimate the pixel-level approximation degree between ground truth (GT) and the prediction.
$\textbf{F-measure}$ ($\textbf{F}_{\beta}$)~\cite{2009-CVPR-Fm} uses the PR information to make a comprehensive analysis.  
The parameter $\beta$ is set to 0.3.
$\textbf{Weighted F-measure}$ ($\textbf{F}^{w}_{\beta}$) , a weighted precision, is designed to improve F-measure.
$\textbf{PR curve}$ compares the prediction results and GT to calculate the precision ($TP/(TP+FP)$) and recall ($TP/(TP+FN)$).
Due to space constraints, we don't show $\textbf{F}_{max}$, $\textbf{F}_{avg}$, $\textbf{E-measure}$, and $\textbf{S-measure}$ in Tab.\ref{T-sota}.

\subsection{Comparision with State-of-the Arts}
For a fair comparison, we use the saliency maps generated by the original codes or provided by the authors.
\begin{figure*}[htb]
\centering
 \includegraphics[width=1.8\columnwidth]{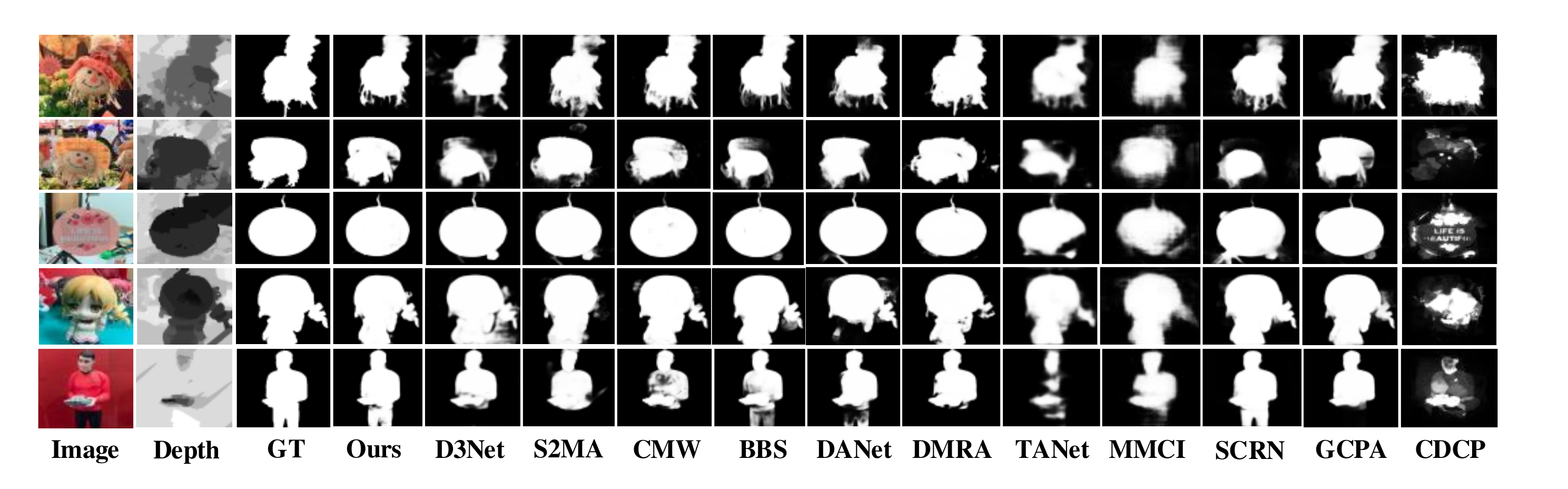} 
\caption{Comparison examples of the GRNet with the state-of-the-art methods. }
\label{sota}
\end{figure*}
We compare GRNet with other 9 methods, including CDCP\cite{2017-ICCVW-CDCP}, D3Net\cite{2020-TNNLS-D3Net}, S2MA\cite{2020-CVPR-S2MA}, DPANet\cite{2020-TIP-DPANet}, CDNet\cite{2021-TIP-CDNet}, cmSalGAN\cite{2021-TMM-cmSalGAN}, SCRN\cite{2019-ICCV-SCRN}, etc.


$\textit{Quantitative Evaluation:}$ 
Some competitive methods are listed in Tab.\ref{T-sota} for $\textbf{F}^{w}_{\beta}$ and $\textbf{MAE}$.  
\textbf{A complete comparison is given in the appendix, including the PR curve.}
Our method performs favorably against others.

$\textit{Qualitative Evaluation:}$ We show the visual comparison in Fig.\ref{sota}. The proposed method makes full use of the advantages of modal complementarity, avoids the interference of invalid information, and produces overall better saliency maps.

\subsection{Ablation Analysis}
1) \textit{Analysis of the contribution of each component:}
Firstly, as shown in the 1st line of Tab.\ref{XR}, we use the ResNet-50 + FPN structure (Fig.\ref{decoder} (b)) as the baseline model, where we only adopt the RGB data.
Then, in the 2nd line, we use two ResNet-50s as multi-modal encoders, use concatenation operation to merge the same level outputs and use FPN as the decoder.
We can find that multi-modal inputs can significantly improve network performance.
In the 4th line, we add MGUs and Recoding Mixer based on the 2nd line setting.
In the 3rd line, we remove the weights ($G_{a}$, $G_{b}$) from MGU relative to the 4th line, which verifies the importance of weight adjustment.
Lines 3 and 4 prove the importance of Recording Mixer to improve network performance, in which MGU plays an important role.
The 5th line uses not only progressive structure (FPN) but also parallel structure.
The 7th adds OEGS (Fig.\ref{decoder} (c)).
The 6th line removes the weight regulation ($G_{r}$, $G_{d}$) obtained by WAM in OEGS.
In the 8th line, we add the well-known IoU loss~\cite{2016-ISVC-IoU} to binary cross entropy loss (BCE).
Structure loss = IoU + BCE.
The WAMs of the above experiments all adopt without MLP design scheme.
In the 9th line, we use the scheme with MLP to enhance the perception ability of WAM.
We can find that every component is essential.
In the last line of Tab.\ref{T-sota}, we use the training strategy of horizontal flip and random crop to further enhance the model GRNet-MLP.

\begin{figure}[t]
\centering
 \includegraphics[width=1\columnwidth]{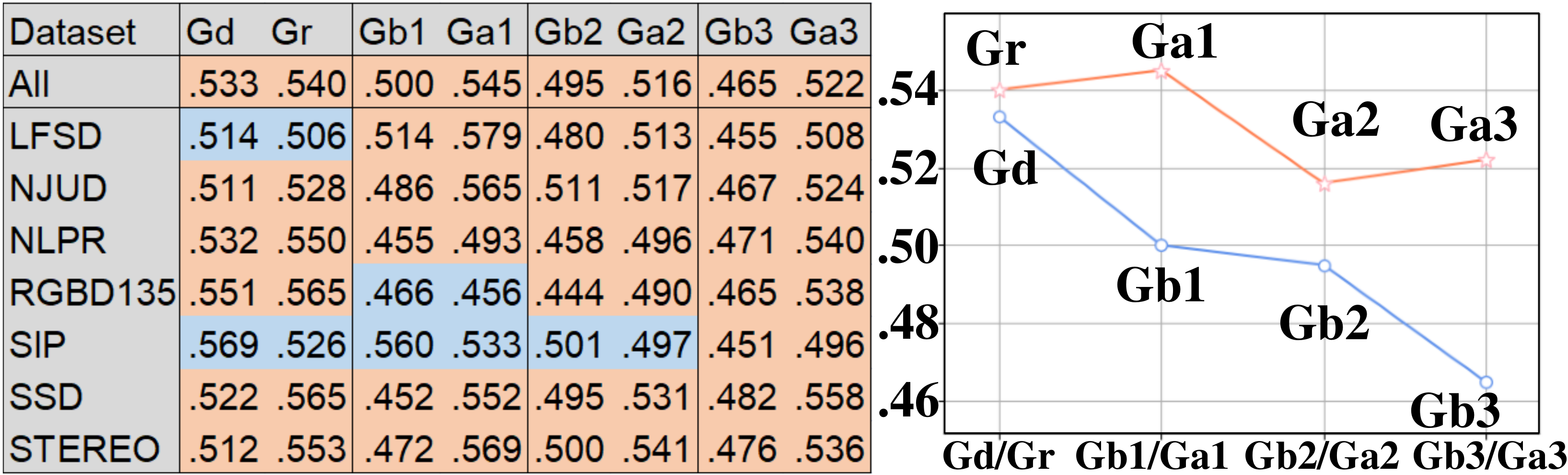} 
\caption{Statistical of analysis of gate units. }
\label{statistic}
\end{figure}

2) \textit{Gate weight analysis:}
In Fig.\ref{statistic}, we analyze the role of weight regulation ($G_{a1}$, $G_{b1}$, $G_{a2}$, $G_{b2}$, $G_{a3}$, $G_{b3}$, $G_{r}$, $G_{d}$) in MGU1, MGU2, MGU3, and WAM of OEGS in a statistical way.
On the left side of the Tab.\ref{statistic}, we show the average weight of each of the seven datasets and the average weight of the combined statistics (ALL).
The blue part indicates that the weight of the depth feature is greater than the weight of the RGB feature ($G_{d}>G_{r}$, $G_{b}>G_{a}$).
We can notice that RGB data is more useful in most cases.
The depth information of some datasets (SIP, LFSD, RGBD135) is of a great reference value.
Besides, through the table and line chart in Fig.\ref{statistic}, we can find that the semantic information (deeper features) of most depth data is relatively insufficient.
While the shallow features of depth data are valuable for detail optimization.

\begin{table}[t]
  \centering
  \caption{Ablation analysis.
  En, Mix, De represent the encoder, mixer, decoder in Fig.\ref{pipline}.
  Mix$^{-}$, De$^{-}$ indicate that the weight adjustment of MGUs and OEGS are removed.}
  \setlength{\tabcolsep}{0.6mm}{
    \begin{tabular}{|l|cc|cc|cc|cc|}
    \hline
    \multicolumn{1}{|c|}{Model} & \multicolumn{2}{c|}{LFSD}             & \multicolumn{2}{c|}{NJU2K}            & \multicolumn{2}{c|}{SIP}              & \multicolumn{2}{c|}{STEREO} \\
                      & F$^{w}_{\beta}\uparrow$               & MAE$\downarrow$                & F$^{w}_{\beta}\uparrow$               & MAE$\downarrow$               & F$^{w}_{\beta}\uparrow$              & MAE$\downarrow$                & F$^{w}_{\beta}\uparrow$               & MAE$\downarrow$  \\
    \hline
    1 w/o depth         & .679              & .126              & .823              & .054              & .783              & .065              & .817              & .050 \\
    2 En+FPN          & .712              & .110              & .842              & .048              & .798              & .059              & .825              & .048 \\
    \hline
    3 En+Mix$^{-}$+FPN         & .743              & .098              & .847              & .047              & .799              & .059              & .831              & .048 \\
    4 En+Mix+FPN          & .748              & .096              & .853              & .045              & .807              & .058              & .833              & .047 \\
    \hline
    5 En+Mix+PF       & .760              & .093              & .853              & .044              & .809              & .057              & .834              & .046 \\
    \hline
    6 En+Mix+De$^{-}$      & .754              & .095              & .854              & .044              & .810              & .056              & .838              & .045 \\
    7 En+Mix+De       & .771              & .088              & .865              & .042              & .813              & .056              & .837              & .045 \\
    \hline
    8 +structure loss & \textcolor[rgb]{ 1,  0,  0}{.808} & \textcolor[rgb]{ 1,  0,  0}{.075} & \textcolor[rgb]{ .357,  .608,  .835}{.882} & \textcolor[rgb]{ .357,  .608,  .835}{.038} & \textcolor[rgb]{ 1,  0,  0}{.841} & \textcolor[rgb]{ 1,  0,  0}{.049} & \textcolor[rgb]{ .357,  .608,  .835}{.869} & \textcolor[rgb]{ .357,  .608,  .835}{.037} \\
    \hline
    9 GRNet+MLP            & \textcolor[rgb]{ .357,  .608,  .835}{.802} & \textcolor[rgb]{ .357,  .608,  .835}{.079} & \textcolor[rgb]{ 1,  0,  0}{.886} & \textcolor[rgb]{ 1,  0,  0}{.036} & \textcolor[rgb]{ .357,  .608,  .835}{.840} & \textcolor[rgb]{ 1,  0,  0}{.049} & \textcolor[rgb]{ 1,  0,  0}{.871} & \textcolor[rgb]{ 1,  0,  0}{.036} \\
    \hline
    \end{tabular}%
    }
  \label{XR}%
\end{table}%

\section{Conclusion}
In this letter, we propose a modal-adaptive gated recoding network for RGB-D salient object detection.
We use the filtering mechanism of the gate unit to reconstruct and recode the features provided by the single-mode encoder.
The optional edge guide stream and hybrid branch decoder can effectively optimize and integrate multi-level multi-modal features.
Extensive evaluation verifies the superiority of the proposed method.
We can find that the depth data plays an auxiliary role from the weight analysis and its shallow features are more valuable, which is helpful for future research.

\bibliographystyle{IEEEtran}
\bibliography{ref}

\begin{thebibliography}{10}
\providecommand{\url}[1]{#1}
\csname url@samestyle\endcsname
\providecommand{\newblock}{\relax}
\providecommand{\bibinfo}[2]{#2}
\providecommand{\BIBentrySTDinterwordspacing}{\spaceskip=0pt\relax}
\providecommand{\BIBentryALTinterwordstretchfactor}{4}
\providecommand{\BIBentryALTinterwordspacing}{\spaceskip=\fontdimen2\font plus
\BIBentryALTinterwordstretchfactor\fontdimen3\font minus
  \fontdimen4\font\relax}
\providecommand{\BIBforeignlanguage}[2]{{%
\expandafter\ifx\csname l@#1\endcsname\relax
\typeout{** WARNING: IEEEtran.bst: No hyphenation pattern has been}%
\typeout{** loaded for the language `#1'. Using the pattern for}%
\typeout{** the default language instead.}%
\else
\language=\csname l@#1\endcsname
\fi
#2}}
\providecommand{\BIBdecl}{\relax}
\BIBdecl

\bibitem{2019-SPL-EACNNet}
W.~Guan, T.~Wang, J.~Qi, L.~Zhang, and H.~Lu, ``Edge-aware convolution neural
  network based salient object detection,'' \emph{IEEE Signal Process. Lett.},
  vol.~26, no.~1, pp. 114--118, 2019.

\bibitem{2018-SPL-BGFANet}
Y.~Zhuge, G.~Yang, P.~Zhang, and H.~Lu, ``Boundary-guided feature aggregation
  network for salient object detection,'' \emph{IEEE Signal Process. Lett.},
  vol.~25, no.~12, pp. 1800--1804, 2018.

\bibitem{2016-SPL-DCMC}
R.~Cong, J.~Lei, C.~Zhang, Q.~Huang, X.~Cao, and C.~Hou, ``Saliency detection
  for stereoscopic images based on depth confidence analysis and multiple cues
  fusion,'' \emph{IEEE Signal Process. Lett.}, vol.~23, no.~6, pp. 819--823,
  2016.

\bibitem{what-is-SOD1}
L.~Itti, C.~Koch, and E.~Niebur, ``A model of saliency-based visual attention
  for rapid scene analysis,'' \emph{IEEE Trans. Pattern Anal. Mach. Intell.},
  vol.~20, no.~11, pp. 1254--1259, 1998.

\bibitem{2009-CVPR-Fm}
R.~Achanta†, S.~Hemami‡, F.~Estrada†, and S.~Su¨sstrunk†,
  ``Frequency-tuned salient region detection,'' in \emph{Proc. IEEE Conf.
  Comput. Vis. Pattern Recog.}, 2009, p. 1597–1604.

\bibitem{2014-ICIMCS-DES}
Y.~{Cheng}, H.~{Fu}, X.~{Wei}, J.~{Xiao}, and X.~{Cao}, ``Depth enhanced
  saliency detection method,'' in \emph{Proc. Int. Conf. Internet Multimedia
  Comput. Service}, 2014.

\bibitem{2017-ICCVW-CDCP}
C.~{Zhu}, G.~{Li}, W.~{Wang}, and R.~{Wang}, ``An innovative salient object
  detection using center-dark channel prior,'' in \emph{Proc. IEEE Int. Conf.
  Comput. Vis. Workshops}, 2017.

\bibitem{2010-ACMTOG-RepFinder}
M.-M. Cheng, F.-L. Zhang, N.~J. Mitra, X.~Huang, and S.-M. Hu, ``Repfinder:
  Finding approximately repeated scene elements for image editing,'' \emph{ACM
  Trans. Graph.}, vol.~29, no.~4, Jul. 2010.

\bibitem{2009-ACMTOG-Sketch2Photo}
T.~Chen, M.-M. Cheng, P.~Tan, A.~Shamir, and S.-M. Hu, ``Sketch2photo: Internet
  image montage,'' \emph{ACM Trans. Graph.}, vol.~28, no.~5, pp. 1--10, Dec.
  2009.

\bibitem{2009-CVPR-tracking-SBDT}
V.~Mahadevan and N.~Vasconcelos, ``Saliency-based discriminant tracking,'' in
  \emph{Proc. IEEE Conf. Comput. Vis. Pattern Recog.}, 2009, pp. 1007--1013.

\bibitem{2020-TIP-DPANet}
Z.~Chen, R.~Cong, Q.~Xu, and Q.~Huang, ``{DPANet}:depth potentiality-aware
  gated attention network for {RGB-D} salient object detection,'' \emph{IEEE
  Trans. Image Process.}, 2020.

\bibitem{1979-TSMC-Ostu}
N.~Otsu, ``A threshold selection method from gray-level histograms,''
  \emph{IEEE Trans. Syst., Man, Cybern.}, vol.~9, no.~1, pp. 62--66, 1979.

\bibitem{2020-ECCV-GateNet}
X.~Zhao, Y.~Pang, L.~Zhang, H.~Lu, and L.~Zhang, ``Suppress and balance: A
  simple gated network for salient object detection,'' in \emph{Proc. Eur.
  Conf. Comput. Vis.}, 2020.

\bibitem{2017-CVPR-FPN}
T.-Y. Lin, P.~Dollár, R.~Girshick, K.~He, B.~Hariharan, and S.~Belongie,
  ``Feature pyramid networks for object detection,'' in \emph{Proc. IEEE Conf.
  Comput. Vis. Pattern Recog.}, 2017, pp. 936--944.

\bibitem{2018-arXiv-BAM}
J.~Park, S.~Woo, J.~Lee, and I.~S. Kweon, ``{BAM:} bottleneck attention
  module,'' in \emph{Proc. British Mach. Vis. Conf.}, 2018, p. 147.

\bibitem{2018-IJCAI-R3Net}
Z.~Deng, X.~Hu, L.~Zhu, X.~Xu, J.~Qin, and G.~Han, ``R$^{3}${N}et: Recurrent
  residual refinement network for saliency detection,'' in \emph{IJCAI Int.
  Joint Conf. Artif. Intell.}, 2018, pp. 684--690.

\bibitem{2019-CVPR-PFANet}
T.~{Zhao} and X.~{Wu}, ``Pyramid feature attention network for saliency
  detection,'' in \emph{Proc. IEEE Conf. Comput. Vis. Pattern Recog.}, 2019,
  pp. 3080--3089.

\bibitem{2018-CVPR-BMPM}
L.~{Zhang}, J.~{Dai}, H.~{Lu}, Y.~{He}, and G.~{Wang}, ``A bi-directional
  message passing model for salient object detection,'' in \emph{Proc. IEEE
  Conf. Comput. Vis. Pattern Recog.}, 2018, pp. 1741--1750.

\bibitem{2018-CVPR-DGRL}
T.~Wang, L.~Zhang, S.~Wang, H.~Lu, G.~Yang, X.~Ruan, and A.~Borji, ``Detect
  globally, refine locally: A novel approach to saliency detection,'' in
  \emph{Proc. IEEE Conf. Comput. Vis. Pattern Recog.}, 2018, pp. 3127--3135.

\bibitem{2018-CVPR-PAGRN}
X.~Zhang, T.~Wang, J.~Qi, H.~Lu, and G.~Wang, ``Progressive attention guided
  recurrent network for salient object detection,'' in \emph{Proc. IEEE Conf.
  Comput. Vis. Pattern Recog.}, 2018, pp. 714--722.

\bibitem{2019-CVPR-PAGE}
W.~Wang, S.~Zhao, J.~Shen, S.~C.~H. Hoi, and A.~Borji, ``Salient object
  detection with pyramid attention and salient edges,'' in \emph{Proc. IEEE
  Conf. Comput. Vis. Pattern Recog.}, 2019, pp. 1448--1457.

\bibitem{2019-ICCV-EGNet}
J.-X. Zhao, J.-J. Liu, D.-P. Fan, Y.~Cao, J.~Yang, and M.-M. Cheng,
  ``{EGNet}:edge guidance network for salient object detection,'' in
  \emph{Proc. IEEE Int. Conf. Comput. Vis.}, Oct 2019.

\bibitem{2019-ICCV-DMRA-DUTRGBD}
Y.~{Piao}, W.~{Ji}, J.~{Li}, M.~{Zhang}, and H.~{Lu}, ``Depth-induced
  multi-scale recurrent attention network for saliency detection,'' in
  \emph{Proc. IEEE Int. Conf. Comput. Vis.}, 2019.

\bibitem{2018-CVPR-PCANet}
H.~{Chen} and Y.~{Li}, ``Progressively complementarity-aware fusion network for
  {RGB}-{D} salient object detection,'' in \emph{Proc. IEEE Conf. Comput. Vis.
  Pattern Recog.}, 2018.

\bibitem{2018-PR-MMCI}
H.~{Chen}, Y.~{Li}, and D.~{Su}, ``Multi-modal fusion network with multi-scale
  multi-path and cross-modal interactions for {RGB}-{D} salient object
  detection,'' \emph{Pattern Recognit.}, 2018.

\bibitem{2017-IROS-M3Net}
H.~{Chen}, Y.-F. {Li}, and D.~{Su}, ``M${3}${N}et: Multi-scale multi-path
  multi-modal fusion network and example application to {RGB}-{D} salient
  object detection,'' in \emph{IEEE Int. Conf. Intell. Rob. Syst.}, 2017.

\bibitem{2014-LNCS-NLPR}
H.~{Peng}, B.~{Li}, W.~{Xiong}, W.~{Hu}, and R.~{Ji}, ``{RGBD} salient object
  detection: A benchmark and algorithms,'' \emph{Proc. Eur. Conf. Comput.
  Vis.}, 2014.

\bibitem{2015-SPIC-NJUD}
R.~{Ju}, Y.~{Liu}, T.~{Ren}, L.~{Ge}, and G.~{Wu}, ``Depth-aware salient object
  detection using anisotropic center-surround difference,'' \emph{Signal
  Process. Image Commun.}, 2015.

\bibitem{2014-ICIMCS-RGBD135}
H.~W. X. X. J. C.~X. Cheng, Yupeng;~Fu, ``Depth enhanced saliency detection
  method,'' \emph{Proc. Int. Conf. Internet Multimedia Comput. Service}, 2014.

\bibitem{2020-TNNLS-SINet}
D.-P. Fan, Z.~Lin, Z.~Zhang, M.~Zhu, and M.-M. Cheng, ``{Rethinking RGB-D
  Salient Object Detection: Models, Datasets, and Large-Scale Benchmarks},''
  \emph{IEEE Trans. Neural Netw. Learn. Syst.}, 2020.

\bibitem{2017-ICCVW-SSD}
{A Three-Pathway Psychobiological Framework of Salient Object Detection Using
  Stereoscopic Technology}, ``Chunbiao {Zhu} and ge {Li},'' in \emph{Proc. IEEE
  Int. Conf. Comput. Vis. Workshops}, 2017.

\bibitem{2012-CVPR-STEREO}
L.~stereopsis for~saliency analysis, ``Yuzhen {Niu} and yujie {Geng} and
  xueqing {Li} and feng {Liu},'' in \emph{Proc. IEEE Conf. Comput. Vis. Pattern
  Recog.}, 2012.

\bibitem{2017-TPAMI-LFSD}
N.~Li, J.~Ye, Y.~Ji, H.~Ling, and J.~Yu, ``Saliency detection on light field,''
  \emph{IEEE Trans. Pattern Anal. Mach. Intell.}, 2017.

\bibitem{2012-CVPR-MAE}
F.~Perazzi, P.~Krahenbuhl, Y.~Pritch, and A.~Hornung, ``Saliency filters:
  Contrast based filtering for salient region detection,'' in \emph{Proc. IEEE
  Conf. Comput. Vis. Pattern Recog.}, 2012, pp. 733--740.

\bibitem{2020-TNNLS-D3Net}
D.-P. Fan, Z.~Lin, Z.~Zhang, M.~Zhu, and M.-M. Cheng, ``{Rethinking RGB-D
  Salient Object Detection: Models, Datasets, and Large-Scale Benchmarks},''
  \emph{IEEE Trans. Neural Netw. Learn. Syst.}, 2020.

\bibitem{2020-CVPR-S2MA}
N.~Liu, N.~Zhang, and J.~Han, ``Learning selective self-mutual attention for
  rgb-d saliency detection,'' in \emph{Proc. IEEE Conf. Comput. Vis. Pattern
  Recog.}, 2020.

\bibitem{2021-TIP-CDNet}
W.-D. Jin, J.~Xu, Q.~Han, Y.~Zhang, and M.-M. Cheng, ``{CDNet}: Complementary
  depth network for {RGB}-{D} salient object detection,'' \emph{IEEE Trans.
  Image Process.}, vol.~30, pp. 3376--3390, 2021.

\bibitem{2021-TMM-cmSalGAN}
B.~Jiang, Z.~Zhou, X.~Wang, J.~Tang, and L.~Bin, ``{cmSalGAN}: {RGB}-{D}
  salient object detection with cross-view generative adversarial networks,''
  \emph{IEEE Trans. Multimedia}, 2020.

\bibitem{2019-ICCV-SCRN}
Z.~Wu, L.~Su, and Q.~Huang, ``Stacked cross refinement network for edge-aware
  salient object detection,'' in \emph{Proc. IEEE Int. Conf. Comput. Vis.}, Oct
  2019.

\bibitem{2016-ISVC-IoU}
M.~A. Rahman and Y.~Wang, ``Optimizing intersection-over-union in deep neural
  networks for image segmentation,'' in \emph{Proc. Int. Symp. Vis. Comput.},
  2016.

\end{thebibliography}

\end{document}